\documentclass{article}

\usepackage{arxiv}

\usepackage[utf8]{inputenc} 
\usepackage[T1]{fontenc}    
\usepackage{hyperref}       
\usepackage{url}            
\usepackage{booktabs}       
\usepackage{amsfonts}       
\usepackage{nicefrac}       
\usepackage{microtype}      
\usepackage{lipsum}		
\usepackage{graphicx}
\usepackage{natbib}
\usepackage{doi}
\usepackage{changes} 
\usepackage{todonotes}
\usepackage{amssymb} 
\usepackage{amsthm}
\usepackage{bm}
\usepackage{amsmath}

\title{A Novel CropdocNet for Automated Potato Late Blight Disease Detection from the Unmanned Aerial Vehicle-based Hyperspectral Imagery}

\author{ {\includegraphics[scale=0.06]{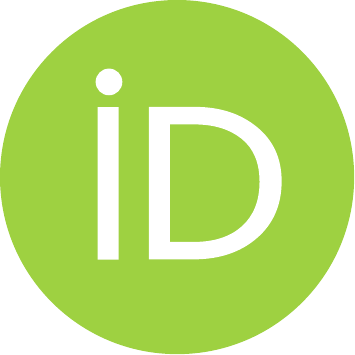}\hspace{1mm}Yue~Shi}\\
	Department of Computing and Mathematics,\\
	Faculty of Science and Engineering,\\
	Manchester Metropolitan University, Manchester M1 5GD, UK. \\
	\texttt{Y.Shi@mmu.ac.uk} \\
	\And
	{\includegraphics[scale=0.06]{orcid.pdf}\hspace{1mm}Liangxiu ~Han} \\
	Department of Computing and Mathematics,\\
	Faculty of Science and Engineering,\\
	Manchester Metropolitan University, Manchester M1 5GD, UK. \\
	\texttt{L.Han@mmu.ac.uk} \\
	\And
	{\includegraphics[scale=0.06]{orcid.pdf}\hspace{1mm}Anthony~Kleerekoper} \\
	Department of Computing and Mathematics,\\
	Faculty of Science and Engineering,\\
	Manchester Metropolitan University, Manchester M1 5GD, UK. \\
	\And
	{\includegraphics[scale=0.06]{orcid.pdf}\hspace{1mm}Sheng~Chang} \\
	Key Laboratory of Digital Earth Science, \\
	Aerospace Information Research Institute, \\
	Chinese Academy of Sciences, Beijing 100094, China \\
	\And
	{\includegraphics[scale=0.06]{orcid.pdf}\hspace{1mm}Tongle~Hu} \\
	College of plant protection, \\
	Hebei Agriculture university, \\
	Hebei, China \\

}

\renewcommand{\shorttitle}{\textit{arXiv} Template}

\hypersetup{
pdftitle={A template for the arxiv style},
pdfsubject={q-bio.NC, q-bio.QM},
pdfauthor={David S.~Hippocampus, Elias D.~Striatum},
pdfkeywords={First keyword, Second keyword, More},
}

\begin{document}
\maketitle

\begin{abstract}
Late blight disease is one of the most destructive diseases in potato crop, leading to serious yield losses globally. Accurate diagnosis of the disease at early stage is critical for precision disease control and management. Current farm practices in crop disease diagnosis are based on manual visual inspection, which is costly, time consuming, subject to individual bias. Recent advances in imaging sensors (e.g. RGB, multiple spectral  and hyperspectral cameras), remote sensing and machine learning offer the opportunity to address this challenge. Particularly,  hyperspectral imagery (HSI) combining with machine learning/deep learning approaches is preferable for accurately identifying specific plant diseases because the HSI consists of a wide range of high-quality reflectance information beyond human vision, capable of capturing both spectral-spatial information. However, due to the complexity and diversity in the reflectance radiation of crop diseases, most of existing machine learning/deep learning based models using HSI imagery are mainly focused on either spectral-based, or spatial-based or spectral-spatial based features but fail to capture the associated hierarchical structure of spectral-spatial attributes of crop diseases (for instance, the part-to-whole relationship between the canopy structural information and reflectance radiation variance of the ground objects hidden in HSI data), which are  important indicators for crop disease diagnosis. \\
Different from the existing methods, this study has proposed a novel deep learning model (CropdocNet) for accurate and automated late blight disease diagnosis. The proposed method considers the potential disease specific reflectance radiation variance caused by the canopy structural diversity, introduces the multiple capsule layers to model the hierarchical structure of the spectral-spatial disease attributes with the encapsulated features to represent the various classes and the rotation invariance of the disease attributes in the feature space. We have evaluated the proposed method with the real UAV-based HSI data under the controlled field conditions. The effectiveness of the hierarchical features has been quantitatively assessed and compared with the existing representative machine learning/deep learning methods. The experiment results show that the proposed model significantly improves the accuracy performance when considering hierarchical-structure of spectral-spatial features, comparing to the existing methods only using spectral, or spatial or spectral-spatial features without consider hierarchical-structure of spectral-spatial features. 
\end{abstract}


\section{Introduction}
\label{sec:intro}
Potato late blight disease, caused by \textit{Phytophthora infestans (Mont.) de Bary}, is one of the most destructive potato diseases, resulting in significant potato yield loss across the major potato growing areas worldwide \cite{demissie2019integrated, patil2017comparision}. The yield loss to the infestation of late blight disease is around $30\%$ to $100\%$ \cite{hirut2017yield, namugga2018yield}. The current control measure mainly relies on application of fungicides \cite{zhang2020integrated}, which is expensive and has negative impacts on the environment and human health due to excessive use of pesticides. Therefore,  early accurate detection of potato late blight disease is vital for effective disease control and management with minimal application of fungicides. \par

Since the late blight disease affects the potato leaves, stems and tubers with visible symptoms (e.g. black lesions with granular regions and green halo) \cite{gao2021automatic, lehsten2017earlier}, the current detection of late blight disease in practice is mainly based on the visual observation \cite{sharma2017image, tung2018combining}. However, this manual inspection method is time consuming and costly, and often causes a delay in the late blight disease management, especially at an early stage across large fields \cite{franceschini2019feasibility}. In addition, the field surveyors diagnose the diseases based on their domain knowledge, which may introduce inconsistency and bias due to individual subjectivity \cite{islam2017detection}. An automated approach for fast and reliable potato late blight disease diagnose is important to ensure effective disease management and control. \par

With the advancements in low-cost sensor technology, computer vision and remote sensing, the image-based agricultural management approaches have shown great potential for automatic crop disease diagnosis based on various types of images (such as the red, green and blue (RGB) images, thermal images, multispectral and hyperspectral images) \cite{dhingra2018study, zhu2018application, shi2018wavelet, shin2021deep, shi2017detection, yang2019tea}. For instance, Shin \textit{et al.} \cite{shin2021deep} employed the RGB images to detect the powdery mildew disease on strawberry leaves through different colour spaces and spatial information. Shi \textit{et al.} \cite{shi2017detection} measured the multispectral reflectance of the wheat yellow rust and aphid to analyse the sensitive spectral bands for the crop pests and diseases, and they proposed a spectral indices based model to detect and discriminate the yellow rust and aphid from the winter wheat leaves. Yang \textit{et al.} \cite{yang2019tea} used the thermal sensor to detect the tea diseases by measuring the canopy water and microclimate associated tea growth status. However, the limited bands used in the RGB, thermal, or multispectral images are always hard to capture the detailed disease features. 

Benefiting from many more narrow spectral bands over a contiguous spectral range, hyperspectral imaging (HSI) provides spatial information in two dimensions and rich spectral information in the third one, which captures detailed spectral-spatial information of the disease infestation with the potential to provide better diagnostic accuracy  \cite{moghadam2017plant, golhani2018review}.  However, how to extract the effective infestation features from the abundant spectral and spatial information from hyperspectral images is a key challenge for disease diagnosis. Currently, based on the features used in the HSI-based disease detection, the existing models can be divided into three categories: \textbf{\textit{spectral feature-based approaches}} focusing on spectral signatures composed by the associated radiation signal of each pixel of image scene in various spectral ranges;  \textbf{\textit{spatial feature-based approaches}} focusing on features such as shape, texture and geometrical structures, and  \textbf{\textit{the joint spectral-spatial feature-based approaches}} focusing on combination of spectral and spatial features.

In the category of \textbf{\textit{spectral feature-based approaches}},  Zhang \textit{et al.}\cite{zhang2019development} proposed a Fusarium head blight (FHB) classification index based on the sensitive spectral features extracted from the HSI to detect the Fusarium head blight with an overall classification accuracy of $89.80\%$. Khan \textit{et al.} \cite{khan2020detection} employed the sub-window permutation analysis (SPA) to extract the wheat powdery mildew associated spectral information. The spectral information are subsequently input into a partial least squares-linear discriminant analysis (PLS-LDA) classifier to detect the diseased wheat leaves from the healthy wheat leaves. Abdulridha \textit{et al.} \cite{abdulridha2019uav} applied classical machine learning methods (KNN and RBF) to HSI images collected from the UAVs for detecting the citrus canker in laboratory conditions. Their findings suggested that the spectral based Modified Chlorophyll Absorption in Reflectance Index (MCARI) is sensitive to the occurrence of the diseases. Nagasubramanian \textit{et al.} \cite{nagasubramanian2018hyperspectral} selected the optimal spectral bands as the input of the Genetic Algorithm (GA) based SVM for early identification of charcoal rot disease in soybean, with a $97\%$ classification accuracy. Huang \textit{et al.} \cite{huang2020diagnosis} extracted twelve sensitive spectral features for Fusarium head blight, which were then fed into a SVM model to diagnose the Fusarium head blight severity with good performance. 

For the category of \textbf{\textit{spatial feature-based approaches}}, the spatial texture of the hyperspectral image represents the foliar contextual variances, such as the colour, density, and leaf angle, which is one of important factors for crop disease diagnosis \cite{gogoi2018remote}. For example, Mahlein \textit{et al.} \cite{mahlein2016plant} summarized the spatial features of the RGB, multi-spectral, and hyperspectral images used in the automatic detection of disease detection. Their study showed that the spatial properties of the crop leaves were affected by leaf chemicals parameters (e.g., pigments, water, sugars, etc.) and light reflected from internal leaf structures. For instance, the spatial texture of the hyperspectral bands from 400 to 700 nm is mainly influenced by foliar content, and the spatial texture of the bands from 700 to 1,100 nm reflects the leaf structure and internal scattering processes. Yuan \textit{et al.} \cite{yuan2016using} introduced the spatial texture of the satellite data into the spatial angle mapper (SAM) to monitor wheat powdery mildew at the regional level. Arivazhagan \textit{et al.} \cite{arivazhagan2013detection} proposed the colour transformation structure as the input feature of the support vector machine (SVM) for automatic detection of plant diseases in large crop fields. Behmann \textit{et al.} \cite{behmann2018spatial} presented a Random Forest algorithm to produce the spatial reference of hyperspectral images of brown rust and septoria tritici blotch on wheat. Castro \textit{et al.} \cite{de2018automatic} proposed an automatic object-based image analysis (OBIA) algorithm, combining the spatial surface information with Random Forest model, for early weed disease mapping within the crop rows, the spatial features of weeds were used as input of the Random Forest (RF) classifier to achieve the rapid and accurate weed mapping.    \par

In the category of \textbf{\textit{the joint spectral-spatial feature-based approaches}}, there are two main strategies for extracting joint spectral-spatial features to represent the characteristics of the crop diseases in HSI data. The first strategy is to extract spatial and spectral features separately and then combine them together based on the 1D or 2D approaches (e.g. feature stacking, convolutional filters, etc). For example, Xie \textit{et al.} \cite{xie2016spectrum} investigated the spectral and spatial features extracted from hyperspectral imagery for detecting early blight disease on eggplant leaves, and they then stacked these features as the input of an AdaBoost models to detect the healthy and infected samples. Rumpf \textit{et al.} \cite{rumpf2010early} extracted a series of spectral and spatial features from the HSI data respectively, and stacked them as input to a SVM classifier to discriminate diseased from non-diseased sugar beet leaves. The second strategy is to jointly extract the correlated spectral-spatial information of the HSI cube through the 3D kernel based approaches. For instance, Zhang \textit{et al.} \cite{zhang2020extraction} extracted the spectral-spatial features based on an edge-preserving filter (EPF), and then fed them into a SVM classifier for automatically extraction of tree crowns damages from the HSI data. In addition, benefiting from the advanced self-learning performance of 3D convolutional kernel, deep convolutional neural networks (DNN)-based approaches have been developed for crop disease diagnosis \cite{qiao2017utilization, shi2021biologically, behmann2015review, saleem2019plant, zhang2019deep}. For instance, Nguyen \textit{et al.} \cite{nguyen2021early} tested the performance of the 2D convolutional neural network (2D-CNN) and 3D convolutional neural network (3D-CNN) for early detection of grapevine viral diseases. Their findings demonstrated that the 3D convolutional filter is able to produce promising results over the 2D convolutional filter from hyperspectral cubes. Karthik \textit{et al.} \cite{karthik2020attention} introduced the 3D convolutional kernels into an attention mechanism-based residual neural network (RNN) for three tomato diseases, i.e. early blight, late blight, and leaf mold, which achieved an overall accuracy of $98\%$. Regarding the depth of 3D convolutional kernel, Suryawati \textit{et al.} \cite{suryawati2018deep} compared the CNN baselines with the depth of 2, 5, and 13 3D-convolutional layers, their findings suggested that the deeper architecture achieved the higher accuracy for the plant disease detection tasks. Francis \textit{et al.} \cite{francis2019disease} explored the effects of the 3D convolutional filters with the size of $3 \times 3$, $5 \times 5$, and $7 \times 7$ on spectral-spatial feature extraction and plant classification and their findings suggested that the small kernel size might perform better on extracting and integrating the joint spectral-spatial features for representing the various plant classes. Similarly, Zhang \textit{et al.} \cite{zhang2019deep} proposed the 3D convolutional block as the basic unit of the the deep neural network (DNN), which showed that the DNN with 4 3D convolutional blocks performed best with the accurate classification and low computing cost. Nagasubramanian \textit{et al.} \cite{nagasubramanian2019plant} developed a 3D deep convolutional neural network (DCNN), with 8 3D convolutional layers, to extract the deep spectral-spatial features for representing the inoculated stem images from the soybean crops. Kumar \textit{et al.} \cite{kumar2019identification} proposed a 3D convolutional neural network (CNN) with 6 3D convolutional layers to extract the spectral-spatial features for various crop diseases. \par 

Despite existing works are encouraging, the existing models do not consider the hierarchical structure of the spectral and spatial information of the crop diseases (for instance, canopy structural information and reflectance radiation variance of the ground objects hidden in HSI data), which are important indicators for crop disease diagnosis. In fact, changes on reflectance due to plant pathogens and plant diseases are highly disease-specific since the optical properties of plant diseases are related to a number of factors such as foliar pathogens, canopy structural information, pigment content, etc.. Thus, these existing methods couldn't represent the various reflectance radiation of the crop disease, which are affected by the particular combination of multiple factors, such as the foliar biophysical variations,  the appearance of typical fungal structures, canopy structural information, from region to region \cite{golhani2018review}. A deeper reason is that the convolutional kernels in the existing CNN methods are independent to each other, which is hard to model the part-to-whole relationship of the spatial-spatial feature and to characterize the complexity and diversity of the potato late blight disease on HSI data \cite{shi2021biologically}.\par  


Therefore, to address the issue above, the hierarchical structure of the spectral-spatial features should be considered in the learning process. In this paper, we propose a novel CropdocNet for automated detection and discrimination of potato late blight disease.  The proposed CropdocNet introduces multiple capsule layers to handle the hierarchical structure of the spectral-spatial features extracted from HSIs and combines the spectral-spatial features to represent the part-to-whole relationship between the deep features and the target classes (i.e. healthy potato and the potato infested with late blight disease). The proposed model is tested and evaluated against real datasets collected from the potato planting area under a controlled field conditions and existing representative machine learning/deep learning models.  

The remaining part of this paper is organized as follow: section \ref{sec2:materials} describes the study area, data collection, and the proposed model, section \ref{sec:3} presents the experimental results, section \ref{sec:4} provides discussions, and section \ref{sec:5} summarizes this work and highlights the future works. \par

\section{Materials and methods} 
\label{sec2:materials}
\subsection{Data acquisition}
\subsubsection{Study site}
\begin{figure}[]   
\centering  
\includegraphics[width=10 cm]{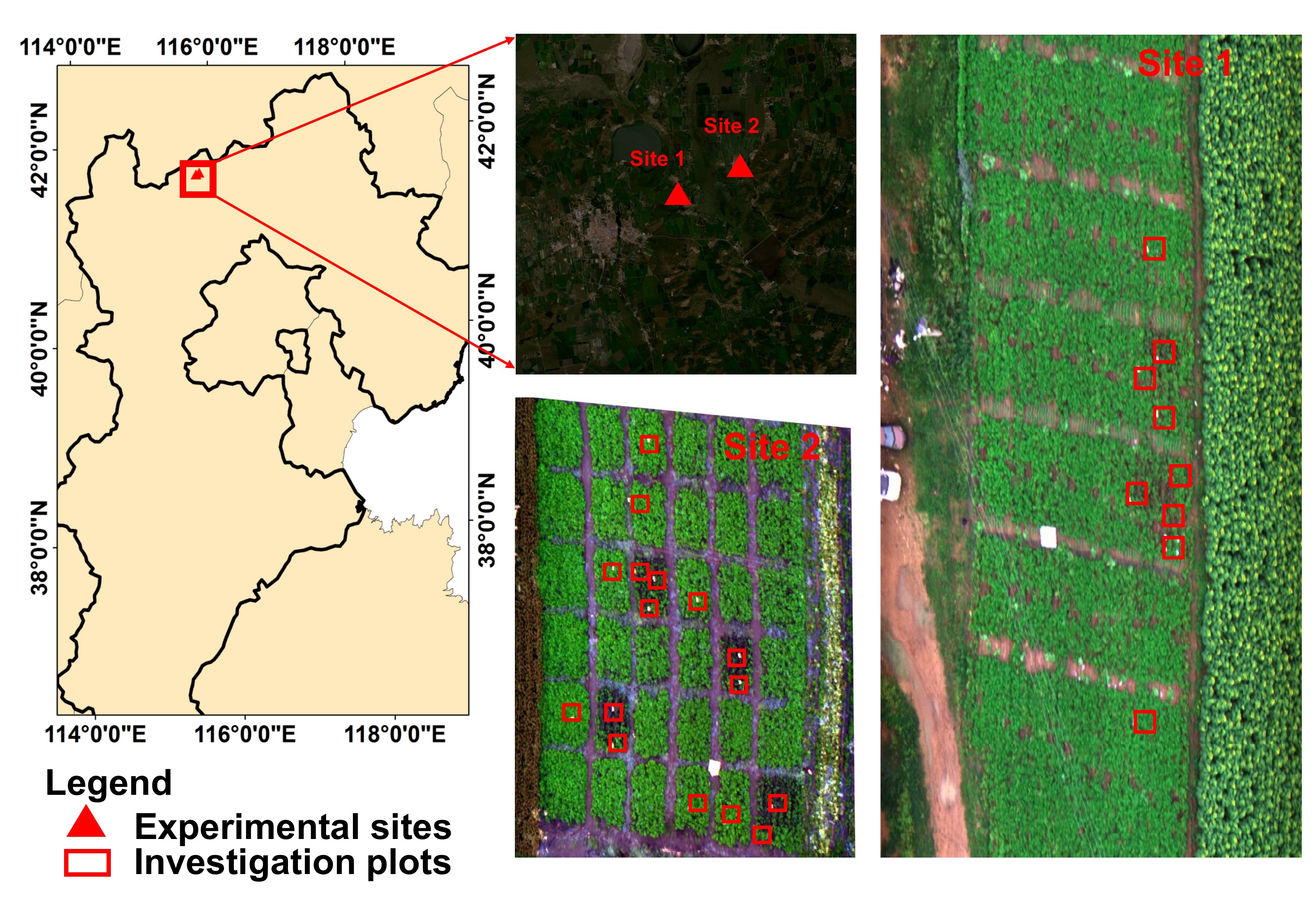}   
\caption{The experimental sites in Guyuan, Hebei province, China.}  
\label{fig:study_site}  
\end{figure}

The field experiments were conducted at two experimental sites in Guyuan, Hebei province, China (see Fig. \ref{fig:study_site}). The site 1 was located at ($41^\circ 41^\prime 2.41^{\prime \prime}$ N, $115^\circ 44^\prime 47.39^{\prime \prime}$ E), where nine $1m \times 1m$ in-situ observation plots were set for the ground truth data investigation. The site 2 was located at ($41^\circ 42^\prime 2.4^{\prime \prime}$ N, $115^\circ 47^\prime 44.39^{\prime \prime}$ E), where a total of eighteen $1m \times 1m$ in-situ observation plots were set for the ground truth data investigation. The field observations were conducted on August 16 and 18, 2020, for experimental site 1 and site 2, respectively. Potato seeding was inoculated on 13 May, 2020. 

\subsubsection{Ground truth disease investigation}
 Four types (classes) of ground truth data were investigated including: healthy potato, late blight disease, soil, and background. Among which, the classes of soil and background can be easily labelled based on visual investigation. For the classes of healthy potato and late blight disease, we firstly investigated the disease ratio of the experiment sits based on National Rules for Investigation and Forecast Technology of the Potato Late Blight ($NY/T1854-2010$). Then, considering the manual bias on the disease investigations, the diseased ratio in a sampling plot lower than $7\%$ was labelled as healthy potato, otherwise it was labelled as diseased potato.

\subsubsection{UAV-based HSIs collection}
The UAV-based HSIs were collected by Dajiang (DJI) S1000 (ShenZhen (SZ) DJI Technology Co Ltd., Gungdong, China) equipped with UHD-185 Imaging spectrometer (Cubert GmbH, Ulm, Baden-Warttemberg, Germany). The collected HSI imagery covering the wavelength ranges from $450$ nm to $950$ nm with 125 bands with a size of $16382 \times 8762 ~pixels$ for experimental site 1, and $8862 \times 7625 ~pixels$ for experimental site 2. All of the UAV-based HSI data were collected at a spatial resolution of 2.5 cm, with a height of 30 m. HSI data were manually labelled based on the ground truth investigations.

\subsection{The proposed CropdocNet model}
Since the traditional convolutional neural networks extract the spectral-spatial features without considering the hierarchical structure representations among the features, it may lead to a poor performance on characterizing the part-to-whole relationship between the features and the target classes. In this study, inspired by the dynamic routing mechanism of capsules \cite{cai2021dynamic}, the proposed CropdocNet model introduces multiple capsule layers (see below) with the aim to model the effective hierarchical structure of spectral-spatial details and generate the encapsulated features to represent the various classes and the rotation invariance of the disease attributes in the feature space for accurate disease detection.  

Essentially, \textbf {the design rationale behind our proposed approach} is that, unlike the traditional CNN methods extracting the abstract scalar features to predict the classes, the spectral-spatial information extracted by the convolutional filters in the form of scalars are encapsulated into a series of hierarchical class-capsules to generate the deep vector features, which represents the specific combination of the spectral-spatial features for the target classes. Based on this rationale, the length of the encapsulated vector features represent the membership degree of an input belonging to a class, and the direction of the encapsulated vector features represent the consistency of the spectral-spatial feature combination between the labelled classes and the predicted classes.

\begin{figure}[]   
\centering  
\includegraphics[width=10 cm]{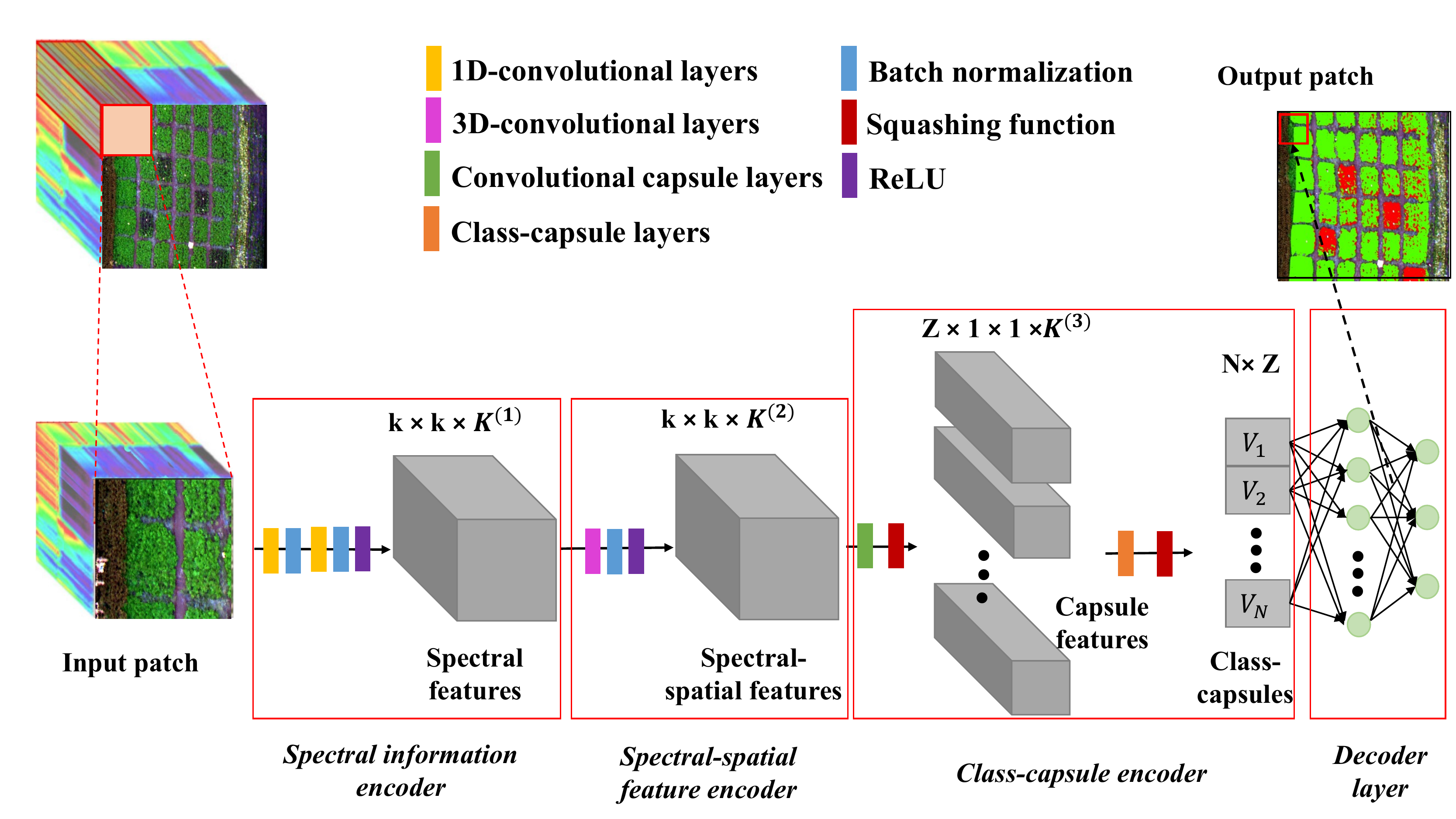}   
\caption{ The workflow of the CropdepcNet framework for potato late blight disease diagnosis }  
\label{fig:workflow}  
\end{figure}

Fig. \ref{fig:workflow} shows the proposed framework, which consists of \textbf{\textit{a spectral information encoder}}, \textbf{\textit{a spectral-spatial feature encoder}}, \textbf{\textit{a class-capsule encoder}}, and \textbf{\textit{a decoder}}. \par

Specifically, the proposed CropdocNet firstly extracts the effective information from the spectral domain based on the 1-D convolutional blocks, and then, it encodes the spectral-spatial details around the central pixels by using the 3-D convolutional blocks. Subsequently, these spectral-spatial features are sent to the hierarchical structure of the class-capsule blocks in order to build the part-to-whole relationship, and to generate the hierarchical vector features for representing the specific classes. Finally, a decoder is employed to predict the classes based on the length and direction of the hierarchical vector features in the feature space. The detailed information for the model blocks are described below. \par

\subsubsection{Spectral information encoder}

The spectral information encoder, located at the beginning of the model, is set to extract the effective spectral information from the input HSI data patches. It is composed of a serial connection of two 1D convolutional layers, two batch normalization layers, and a ReLu layers. \par
Specifically, as shown in Fig. \ref{fig:workflow}, the HSI data with $H$ rows, $W$  columns, and $B$ bands, denoted as $X \in \mathbb{R}^{H \times W \times B}$, can be viewed as a sample set with $H \times W$ pixel vectors. Each of the pixels represents a class. And then, the 3-D patches with a size of $ d \times d \times B $ around each pixel are extracted as the model input, where $d$ is the patch size. In this study, $d$ is set as 13. These patches are labelled with the classes same as their central pixels. \par

Subsequently, the joint 1D convolution and batch normalization series, which receive the data patch from the input HSI cube, are introduced to extract the radiation magnitude of the central band and their neighbour bands. A total of $K^{(1)}$ convolutional kernels with a size of $ 1 \times 1 \times L_{rf} $ are employed by the 1D convolutional layer, where, $L_{rf}$ is the length of the receptive field for the spectral domain. The 1D convolutional layer is calculated as follows: \par

\begin{equation}
C^j_p=\sum_{l=1}^{L_{rf}} W_{l}^j I_{l}^{p}
\end{equation}

where $C_p^j$ is the intermediate output of the $p^{th}$ neuron with the $j^{th}$ kernel, $W_{l}^j$ is the weight for the $l^{th}$ unit of the $j^{th}$ kernel, and $I_{l}^p$ is the feature value of the $l^{th}$ unit corresponding to the $p^{th}$ neuron. \par
The second 1D convolution and batch normalization series are used to extract the abstract spectral details from the low-level spectral features. Finally, a ReLu activation function is used to obtain a spectral feature output denoted as  $\mathbf{X}_{out}^1 \in \mathbb{R}^{H \times W \times K^{(1)}}$. \par

\subsubsection{Spectral-spatial feature encoder}
The spectral-spatial feature encoder is located after the spectral information encoder, which aims to arrange the extracted spectral features in $\mathbf{X}_{out}^1$ into the joint spectral-spatial features that feed to the subsequent capsule encoder. Firstly, a total of $K^{(2)}$ 3D convolutional layer is used on the $\mathbf{X}_{out}^1$ with a kernel size of $c \times c \times K^{(1)}$, where $c$ is the kernel size, which is set as 13 in this study. Then, the a batch normalization step and a ReLu activation function is used to generate the output volume $\mathbf{X}_{out}^2 \in \mathbb{R}^{H \times W \times K^{(2)}}$. \par

\subsubsection{Class-capsule encoder}
In this model, a class-capsule encoder is introduced to enhance the hierarchical pose (translational and rotational) correlations between the low-level spectral-spatial information from HSI, and detect the pose and orientation characteristics associated instantiation parameters for the target classes of healthy and diseased potato. It comprises two layers: a feature encapsulation and a class-capsule. \par
The feature encapsulation layer consists of $Z$ convolutional-based capsule units, and each of the capsule unit composed by $K$ convolutional filters, and the size of each filter is $k \times k \times K^{(3)}$. Specifically, for this layer, the $\mathbf{X}_{out}^2$ from the spectral-spatial feature encoder will input into a series of capsules units to learn the potential translational and rotational structure between the features in $\mathbf{X}_{out}^2$. An output vector $\mathbf{u^{(m)}} \in \mathbb{R}^{K} = \boldsymbol{[} u^{(m)}_1,u^{(m)}_2,\cdots{},u^{(m)}_K \boldsymbol{]} $ would be generated by the $K$ convolutional kernel of $m^{th}$ capsule. The orientation of the output vector represents the class-specific hierarchical structure characteristics, while its length represents the degree a capsule is corresponding to a class (e.g. health or disease). To measure the length of the output vector as a probability value, a nonlinear squash function is used as follow:

\begin{equation}
\breve u_m=\frac{||u_m||^2}{1+||u_m||^2}\cdot\frac{u_m}{||u_m||}
\end{equation}
wherein, $\breve u_m^{(l)}$ is the scaled vector of $\mathbf{X}_{out}^2$. This function compresses the short vector features to zero and enlarge the long vector features a value close to 1. The final output is denoted as $\mathbf{X}_{out}^3 \in \mathbb{R}^{ Z \times 1 \times 1 \times K}$.  \par

Subsequently, the class-capsule layer is introduced to encode the encapsulated vector features in $\mathbf{X}_{out}^3$ in to the class-capsule vectors corresponding to the target classes. The length of the class-capsule vectors indicate the probability of belonging to corresponding classes. Here, a dynamic routing algorithm is introduced to iteratively update the parameters between the class-capsule vectors with the previous capsule vectors. The dynamic routing algorithm provides a well-designed learning mechanism between the feature vectors, which reinforces the connection coefficients between the layers, and highlights the part-to-whole correlation relationship between the generated capsule features. Mathematically, the class-capsule $\hat{u}_{n|m}^{(l)}$ is calculated as:
\begin{equation}
\hat{u}_{n|m}^{(l)}=W_{m,n}^{(l)}\cdot\breve{u}_m^{(l-1)}+B_n^{(l)}
\end{equation}
where $W_{m,n}^{(l)}$ is a transformation matrix connecting the layer $l-1$ with layer $l$, $\breve{u}_m^{(l-1)}$ is the $m^{th}$ feature of layer $l-1$, and $B_n^{(l)}$ is the biases. This function allows the vector features in low-level to make prediction for the rotation invariance of high-level features corresponding to the target classes. After that, the prediction agreement can be computed by a dynamic routing coefficient $c_{m,n}^{(l)}$: 

\begin{equation}
s_n^{(l)}=\sum_m^{z^{(l-1)}}c_{m,n}^{(l)}\cdot\hat{u}_{n|m}^{(l)}
\end{equation}

where $c_{m,n}^{(l)}$ is a dynamic routing coefficient measuring the weight of the $m^{th}$ capsule feature of layer $l-1$ activating the $n^{th}$ class-capsule of layer $l$, the sum of all the coefficients would be 1, and the dynamic routing coefficient can be calculated as:
\begin{equation}
c_{m,n}^{(l)}=\frac{e^{b_{m,n}}}{\sum_i^{z^{(l)}}e^{b_{m,i}}}
\end{equation}
where $b_{m,n}$ is the $log$ prior representing the correlation between layer $l-1$ and layer $l$, which is initialized as 0 and is iteratively updated as follow:
\begin{equation}
b_{m,n}^l=b_{m,n}^{l-1}+v_n^{l-1}\cdot\hat{u}_{n|m}^{(l-1)}
\end{equation}
where $v_n^{l}$ is the activated capsule of layer $l$, which can be calculated based on the function as follows: 
\begin{equation}
v_n^{l}=\frac{||s_n^{(l)}||^2}{1+||s_n^{(l)}||^2}\cdot\frac{s_n^{(l)}}{||s_n^{(l)}||}
\end{equation}
Updated by the dynamic routing algorithm, the capsule features with similar prediction will be clustered, and a robust prediction based on these capsule clusters is performed. Finally, the the loss function ($L$) is defined as follow:  
\begin{equation} \label{eq:Lm}
\begin{split}
L_{margin}=\sum_i^{n_{class}}T_i \max{(0,edge^+-||v_n^{l}||^2)}+\\ \mu(1-T_i)(\max(0,||v_n^{l}||-edge^-)^2)
\end{split}
\end{equation}
where $T_i$ is set as 1 when class $i$ is currently classified in the data, otherwise is 0. The $edge^+$, set as 0.9, and $edge^-$, set as 0.1, are defined to force the $v_n^{l}$ into a series of small interval values to update the loss function. $\mu$, defined as 0.5, is a regularization parameter to avoid over-fitting and reduce the effect of the negative activity vectors. \par

\subsubsection{The decoder layer}
The decoder layer, composed by two fully connected layers, is designed to reconstruct the classification map from the output vector features. The final output of this model is regarded as $\tilde{\mathbf{Y}}\in\mathbb{R}^{H \times W}$. For the model updating, the model loss aims to minimize the difference between the labelled map, $\bar{\mathbf{Y}}$, and the output map, $\tilde{\mathbf{Y}}$. The final loss function is defined as follow:
\begin{equation}
L_{end}=L_{margin}+\theta\cdot L_{reconstruction}
\end{equation}

where, $L_{reconstruction}=\lVert{\tilde{\mathbf{Y}}-\bar{\mathbf{Y}}\rVert}$ is the mean square error (MSE) loss between the labelled map and the output map, and $\theta$ is the learning rate, in this study, $\theta$ is set to 0.0005 in order to trade-off the contribution of $L_{margin}$ and $L_{reconstruction}$. And an Adam optimizer is used to optimize the learning process. \par

\subsection{Model evaluation on the detection of potato late blight disease}
\subsubsection{Experimental design}
In order to evaluate the performance of the proposed CropdocNet on the detection of potato late blight disease, three experiments have been conducted: 1) Model sensitivity to the network depth. 2) An accuracy comparison study between the CropdocNet and the existing machine/deep learning models for potato late blight disease detection 3) The accuracy performance evaluation at both pixel and patch scales. The detailed experimental settings are described as follow.\\
 
1) Experiment one: Model sensitivity to the depth of the network  \\

The depth of the network is an important parameter that determines the model performance on spectral-spatial feature extraction. To  investigate the effect of the depth of the network, we change the number of the 1D convolutional layers and the 3D convolutional layers in the proposed model to control the model depth. For each of the configuration, we compare the model performance on the potato late blight disease detection and show the best accuracy. \\

2) Experiment two: An accuracy comparison study between the CropdocNet and the existing machine/deep learning models \\

In order to evaluate the effectiveness of the hierarchical structure of the spectral-spatial information in our model on the detection of potato late blight disease, we compare the proposed CropdocNet considering the hierarchical structure of the spectral-spatial information with the existing representative machine/deep learning approaches using a) spectral features only, b) the spatial features only, and c) joint spectral-spatial features only.   Based on literature review,  SVM, Random Forest (RF) and 3D-CNN are selected as existing representative machine learning/deep learning models for comparison study.  For the spectral feature-based models, the works in \cite{zhao2020identification, nagasubramanian2018hyperspectral, huang2020diagnosis} have reported the Support Vectors Machine (SVM) is an effective classifier for plant disease diagnosis based on spectral feastures. For the spatial feature based models, researches in \cite{behmann2018spatial, de2018automatic, golhani2018review} have demonstrated that the Random Forest (RF) is an effective classifier for analysis of the plant stress associated spatial information in disease diagnosis. For joint spectral-spatial feature based models, a number of deep learning models have been proposed for extracting the spectral-spatial features from the HSI data. Among which, 3D convolutional neural network (3DCNN)-based models \cite{nagasubramanian2019plant, zhang2019deep, francis2019disease} are the most commonly used in plant disease detection. All these existing methods didn't consider the hierarchical structure of the spectral-spatial information. \\ 

3) Experiment three: The accuracy performance evaluation at both pixel and patch scales  \\

 To evaluate the model performance on mapping of the potato blight disease occurrence situation under different observation scales, two evaluation methods are conducted: 1) pixel-scale evaluation, which focuses on the performance evaluation of the proposed model for detection of the detailed late blight disease occurrence at the pixel-level based on the pixel-wised ground truth data.  2) patch-scale evaluation, which focuses on performance evaluation at the patch level by aggregation of the pixel-wised classification into the patches with a given size. For instance, in our case, the field is divided into the $1m \times 1m$ patches/grids, the disease predictions at the pixel level are aggregated into the $1m \times 1m$ patches, which is compared against the corresponding real disease occurrence within that given patch area. This patch-scale evaluation further indicates the classification robustness of the disease detection at different observation scales.    

\subsubsection{Evaluation metrics}
A set of widely used evaluation metrics are introduced to evaluate the accuracy of the detection of potato late blight disease including: Confusion Matrix, Sensitivity, Specificity, Overall Accuracy (OA), Average Accuracy (AA), and Kappa coefficient. The definition of the metrics are set as follow:  \par

\begin{table}[]
\caption{The definition of Confusion Matrix: P = Condition Positive; N = Condition Negative; TP = True Positive; FP = False Positive; TN = True Negative; FN = False Negative; UA = user's accuracy; PA = producer's accuracy. Wherein, the producer's accuracy refers to the probability that a certain class is classified correctly, and the user's accuracy refers to the reliability of a certain class.} 

\label{table:cm_def}
\centering
\resizebox{10 cm}{!}{
\begin{tabular}{cccc}
\toprule
      & P      & N      &  UA (\%)\\    \midrule
P     & TP     & FP     &  $TP/(TP+FP) \times 100\%$ \\
N     & FN     & TN     &  $TN/(TN+FN) \times 100\%$ \\
PA(\%) & $TP/(TP+FN) \times 100\%$ & $TN/(TN+FP) \times 100\%$ &  \\\bottomrule
\end{tabular}
}
\end{table}

\begin{equation}
Sensitivity=TP/{TP+FN}
\end{equation}
    
\begin{equation}
Specificity=TN/{TN+FP}
\end{equation}

\begin{equation}
OA={TP+TN}/{TP+TN+FP+FN}
\end{equation}
    
\begin{equation}
AA=1/2 \times (\frac{TP}{TP+FN}+\frac{TN}{TN+FP})
\end{equation}

\begin{equation}
Observation = TP+TN
\end{equation}

\begin{equation}
Expect = \frac{((TP+FP)*(TP+FN)+(TN+FP)*(TN+FN))}{(TP+TN+FP+FN)}
\end{equation}

\begin{equation}
Kappa = \frac{Observation - Expect}{(TP+TN+FP+FN) - Expect}
\end{equation}

\subsubsection{Model training}
In this study, a total of $1,200$ 3D blocks with a size of $64 \times 64 \times 270$ are extracted as the input patches from the HSI data covering the study area. A total of $1,000$ blocks are randomly selected for 5-fold cross validation, and the remaining $200$ blocks are used as the independent test dataset. For model optimization, an Adam optimizer, with a batch size of $64$, is used to train the proposed model. The learning rate is initially set as $1 \times 10^{-3}$, and is iteratively increased with a step of $1 \times 10^{(-6)}$. 

The hardware environment for model training consists of an Intel (R) Xeon (R) CPU E5-2650, NVIDIA TITAN $X$ (Pascal) and 64 GB memory. The software environment is Tensorflow 2.2.0 framework and python 3.5.2 as the programming language. 

\section{Results}
\label{sec:3}
\subsection{The CropdocNet model sensitivity to the depth of the convolutional filters}
\begin{figure}[]   
\centering  
\includegraphics[width=10.5 cm]{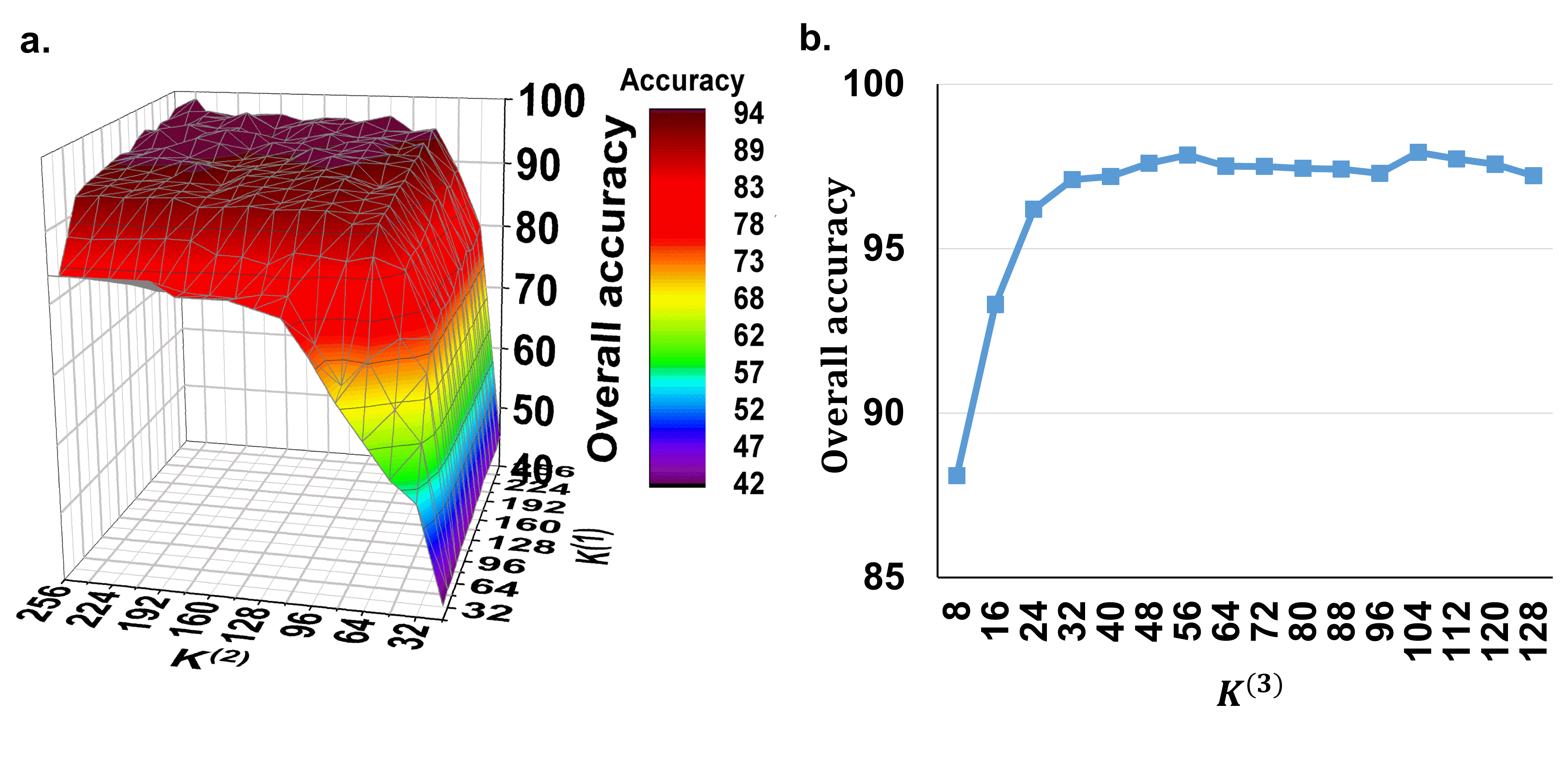}   %
\caption{ The model sensitivity to the depth of the convolutional filters. (a) the overall accuracy of using the different $K^{(1)}$ and $K^{(2)}$ with fixed $K^{(3)}$ of 16. (b) the overall accuracy of using different $K^{(3)}$ under the fixed $K^{(1)}$ and $K^{(2)}$ values of $128$ and $64$. Here, $K^{(1)}$ is the depth of the 1D convolutional layers for the spectral feature extraction, $K^{(2)}$ is the depth of the 3D convolutional layers for the spectral-spatial feature extraction, and $K^{(3)}$ is the number of the capsules vector features.}  
\label{fig:parameters}  %
\end{figure}

In the proposed method, we will need to set the parameters $K^{(1)}$, $K^{(2)}$, and $K^{(3)}$, which represent the depth of the 1D convolutional layers for the spectral feature extraction, the depth of the 3D convolutional layers for the spectral-spatial feature extraction, and the number of the capsules vector features respectively. Due to the fact that, in our model, the high-level capsule vector features are derived from the low-level spectral-spatial scalar features, the depth of the convolutional filters is the main factor that influences this process. Therefore, we firstly set the $K^{(3)}$ as a fixed value of $16$ to evaluate the effect of using different depths of $K^{(1)}$ and $K^{(2)}$ for spectral-spatial scalar feature extraction. Fig. \ref{fig:parameters}a shows the overall accuracy of the potato late blight disease classification using the the various $K^{(1)}$ and $K^{(2)}$ from $32$ to $256$ with a step of $16$. It can be seen that both $K^{(1)}$ and $K^{(2)}$ have the positive effects on the classification accuracy. The accuracy convergence is more sensitive to $K^{(2)}$than to $K^{(1)}$. This is because the $K^{(2)}$ controls the joint spectral-spatial features with more correlation with the plant stress, and affects the final disease recognition accuracy. Overall, the classification accuracy reaches convergence (approximately $85.05\%$) when $K^{(1)} = 128$ and $K^{(2)} = 64$. Thus, in the following experiments, we set $K^{(1)} = 128$ and $K^{(2)} = 64$ for optimal model performance and computing efficiency. \par

Subsequently, we test the effect of the parameter $K^{(3)}$ with the fixed $K^{(1)}$ and $K^{(2)}$ values of $128$ and $64$. Fig. \ref{fig:parameters}b shows that the classification accuracy increases when $K^{(3)}$ increases from $8$ to $32$, and then converges to approximately $97.15\%$ when $K^{(3)}$ is greater than $32$. These findings suggest that the number of $32$ capsule vector blocks is the minimum configuration for our model on detection of potato late blight disease. Therefore, in order to trade off between the model performance and computing performance, $K^{(3)}$ is set as $32$ in the subsequent experiments.  \par

\subsection{Accuracy comparison study between the CropdocNet and existing machine learning-based approaches for potato disease diagnosis}
\label{sec:benchmark}
In this experiment, we quantitatively investigate the performance of the proposed model considering the hierarchical structure of the spectral-spatial information and the representative machine/deep learning approaches without considering it  (i.e. SVM with the spectral feature only, RF with the spatial feature only, and 3D-CNN with the joint spectral-spatial feature only) on potato late blight disease detection with different feature extraction strategies. Wherein, for SVM, we used Radial Basis Function (RBF) kernel to learn the non-linear classifier, two kernel parameters C and $\gamma$ were set to 1000 and 1, respectively \cite{nagasubramanian2018hyperspectral, huang2020diagnosis}. For RF, a quantity of 500 decision trees were employed because this value has been proven to be effective in crop disease detection tasks \cite{de2018automatic, behmann2018spatial}. For 3D-CNN, we employed the model architecture and configurations reported in Nagasubramanian \textit{et al.} \cite{nagasubramanian2019plant}'s study. All of the models were trained on the training dataset and validated on both of the testing and independent datasets.  \\

Table \ref{table:OA} shows the accuracy comparison between the proposed model and the competitors using the test dataset and the independent dataset. The results suggest that the proposed model using the hierarchical vector features consistently outperforms the representative machine/deep learning approaches with scalar features in all of the classes. The OA and AA of the proposed model are $97.33 \%$ and $98.09\%$ respectively with a Kappa value of $0.82$ on the test dataset, which is $7.8\%$ in average higher than the second best model (i.e. the 3D-CNN model with joint spectral-spatial scalar features). In addition, the classification accuracy of the proposed model achieves $96.14\%$, which is $11.8\%$ higher than the second best model. For the independent test dataset, the OA and AA of the proposed model achieve $95.31 \%$ and $95.73\%$ respectively with a Kappa value of $0.80$, which is the best classifier. The classification accuracy achieves $93.36\%$, $9.88\%$ which is higher than the second best model. These findings demonstrate that the proposed model with the hierarchical structure of the spectral-spatial information outperforms scalar spectral-spatial feature based models on the classification accuracy of the late blight disease detection.

To further explore the classification difference significance between the proposed method and the existing machine models, the McNemar’s Chi-Squared ($\chi^{2}$) test is conducted between two-paired models. The statistic significant is shown in Table. \ref{table:class_sig}.  Our results show that the overall accuracy improvement of the proposed model is statistically significant with $\chi^{2} = 32.92 (p \le 0.01)$ for SVM, $\chi^{2} = 31.52 (p \le 0.01)$ for RF, and $\chi^{2} = 29.34 (p \le 0.01)$ for 3D-CNN. 

\begin{figure}[]   
\centering  
\includegraphics[width=10 cm]{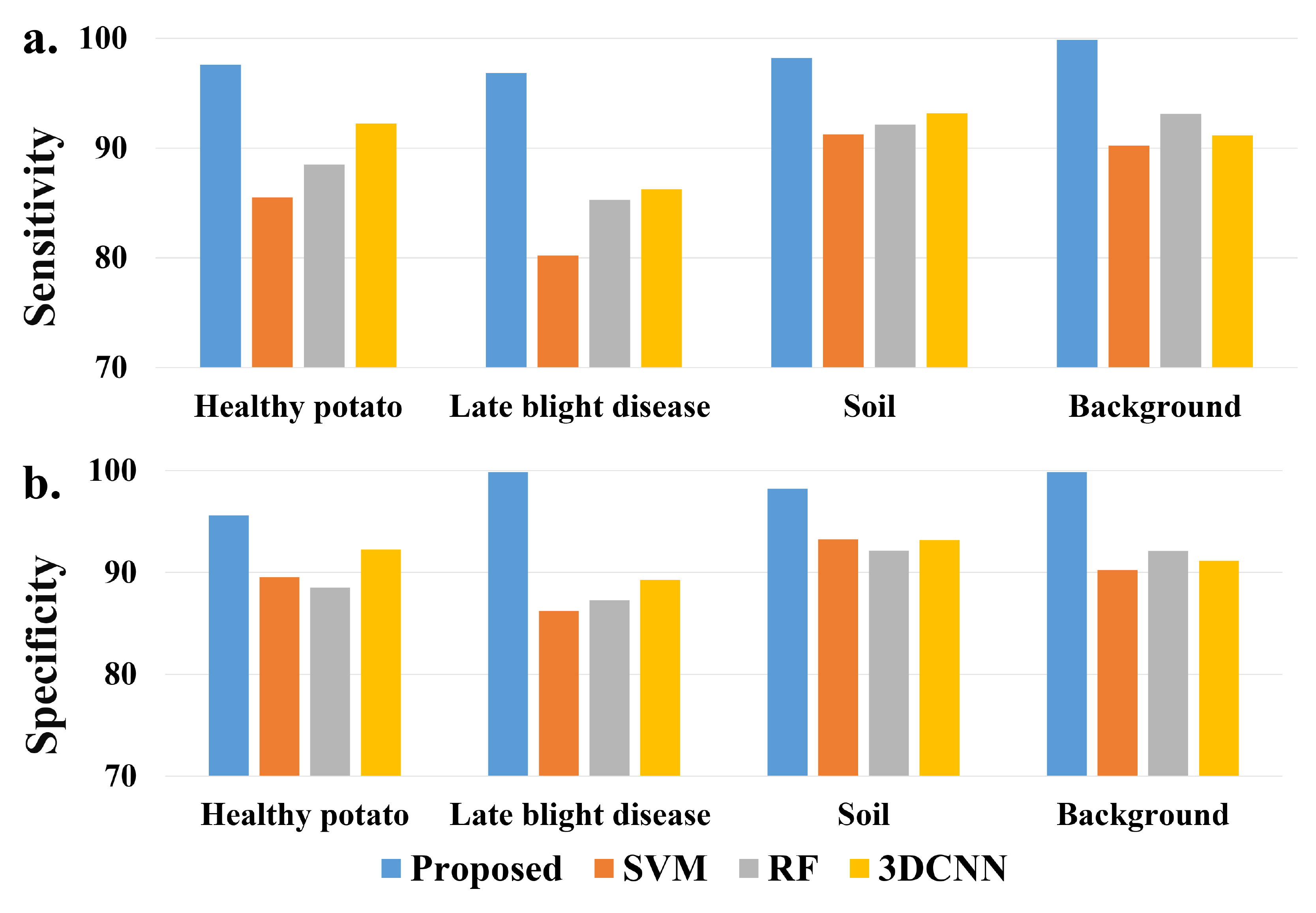}   
\caption{ A comparison of (a) sensitivity and (b) specificity of each classes from different models.}  
\label{fig:sensitivity}  
\end{figure}

Moreover, a sensitivity and specificity comparison of detailed class are provided in Fig. \ref{fig:sensitivity}. Similar to the classification evaluation results, the proposed model achieves the best sensitivity and specificity on all of the ground classes, especially for the class of potato late blight disease.

\begin{table}[]
\caption{The accuracy comparison between the proposed model and existing representative machine/deep learning models on potato late blight disease detection.} 
\label{table:OA}
\centering
\resizebox{10 cm}{!}{
\begin{tabular}{ccccccccc}
\toprule
                    & \multicolumn{4}{c}{Models on test dataset} & \multicolumn{4}{c}{Models on independent test dataset} \\ \cline{2-9}  
Class               & Proposed           & SVM    & RF     & 3D-CNN & Proposed              & SVM       & RF        & 3D-CNN     \\ \midrule
Healthy potato      & \textbf{97.21}     & 86.82  & 90.64  & 94.24  & \textbf{96.32}         & 82.25     & 88.92     & 85.21      \\
Late blight disease & \textbf{96.14}     & 80.15  & 82.31  & 85.51  & \textbf{93.36}         & 71.76     & 79.01     & 83.48      \\
Soil                & \textbf{99.85}     & 89.91  & 92.19  & 93.31  & \textbf{98.44}         & 87.42     & 83.78     & 85.12      \\
Background          & \textbf{99.14}     & 90.31  & 93.52  & 91.16  & \textbf{94.88}        & 89.85     & 86.35     & 83.85      \\
OA(\%)              & \textbf{97.33}     & 84.89  & 87.77  & 90.32  & \textbf{95.31}         & 79.45     & 83.97     & 90.32      \\
AA(\%)              & \textbf{98.09}     & 86.8   & 89.67  & 91.06  & \textbf{95.75}         & 82.82     & 84.52     & 91.06      \\
Kappa               & \textbf{0.822}     & 0.549  & 0.614  & 0.728  & \textbf{0.801}         & 0.512     & 0.595     & 0.699      \\ \bottomrule
\end{tabular}
}
\end{table}

\begin{table}[]
\caption{The McNemar’s Chi-Square Test of the proposed model and the existing representative machine/deep learning models on potato late blight disease detection.} 
\label{table:class_sig}
\centering
\resizebox{10 cm}{!}{
\begin{tabular}{cccccccc}
\toprule
Class               & Proposed vs. SVM & Proposed vs. RF & Proposed vs. 3D-CNN \\ \midrule
Healthy potato      & 31.82**            & 30.25**           & 28.82**               \\
Late blight disease & 35.91**            & 33.24**           & 32.31**               \\
Soil                & 33.25**            & 32.12**           & 30.33**               \\
Background          & 32.15**            & 30.14**           & 27.42**               \\ \hline
Overall             & 32.92**            & 31.52**           & 29.34**               \\ \bottomrule
\end{tabular}
}
\end{table}

\subsection{The model performance on mapping the potato late blight disease from the UAV HSI data}


Table \ref{table:confuse} shows the confusion matrix of the proposed model on the pixel-scale disease classification by using the evaluation dataset of site 1 and site 2. Our results demonstrate that, compared with the accuracies based on the test dataset mentioned in section \ref{sec:benchmark}, the proposed model performs a robust classification on the evaluation dataset with the overall accuracy of $96.5\%$ and Kappa of 0.82. The misclassification mainly occurs between the classes of healthy potato and potato late blight disease, and soil and potato late blight disease. \par
\begin{figure}[]   
\centering  
\includegraphics[width=10.5 cm]{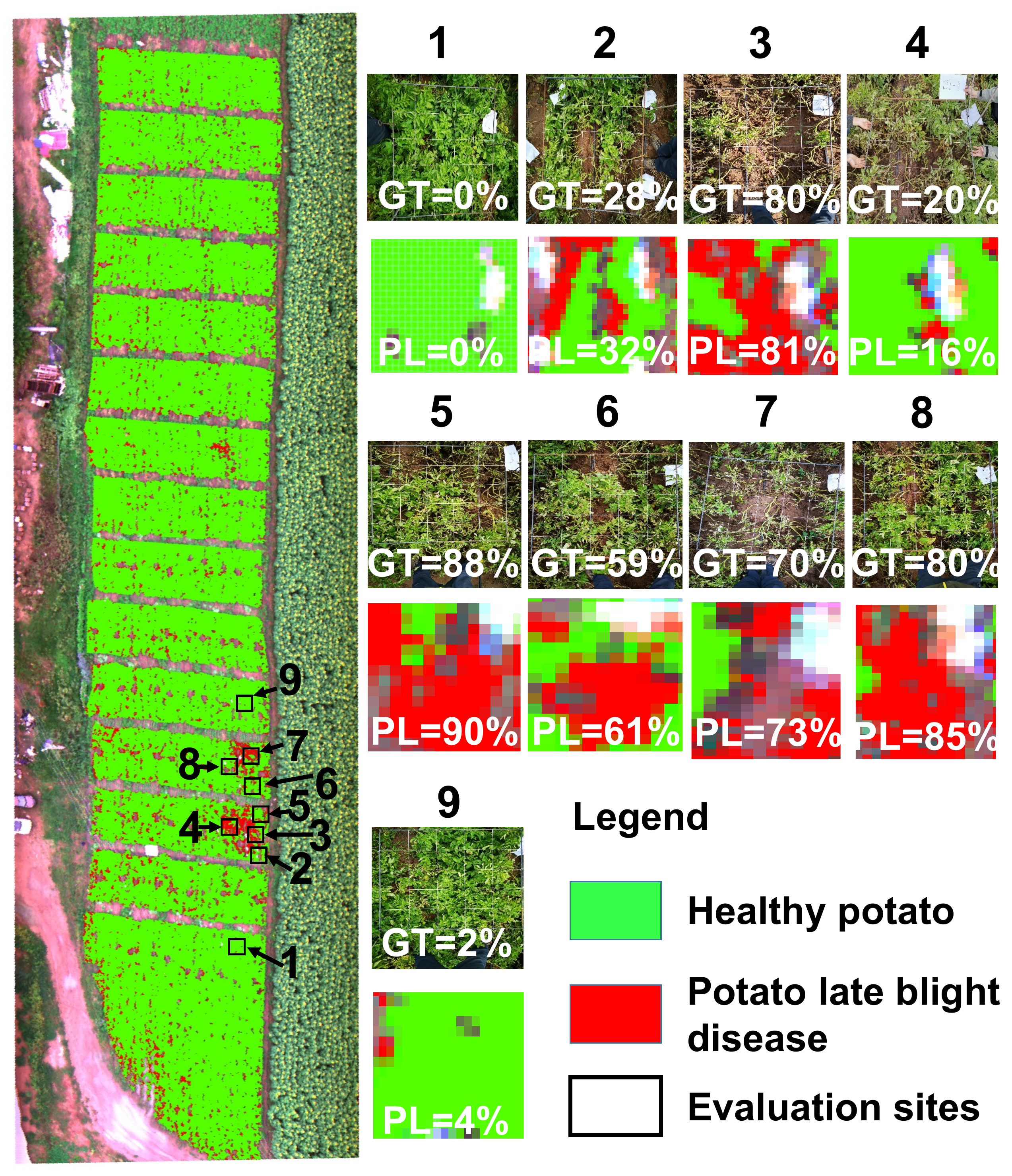}   
\caption{The patch scale test for the classification maps of the healthy potato and the potato late blight disease in experimental site 1. Here, the example patches on the right side illustrate the accuracy comparison between the ground truth (GT) investigations and the predicted levels (PL) of the late blight disease. Each value inside the patch represents the disease ratio.  }  
\label{fig:class_map1}  
\end{figure}
\begin{figure}[]   
\centering  
\includegraphics[width=10.5 cm]{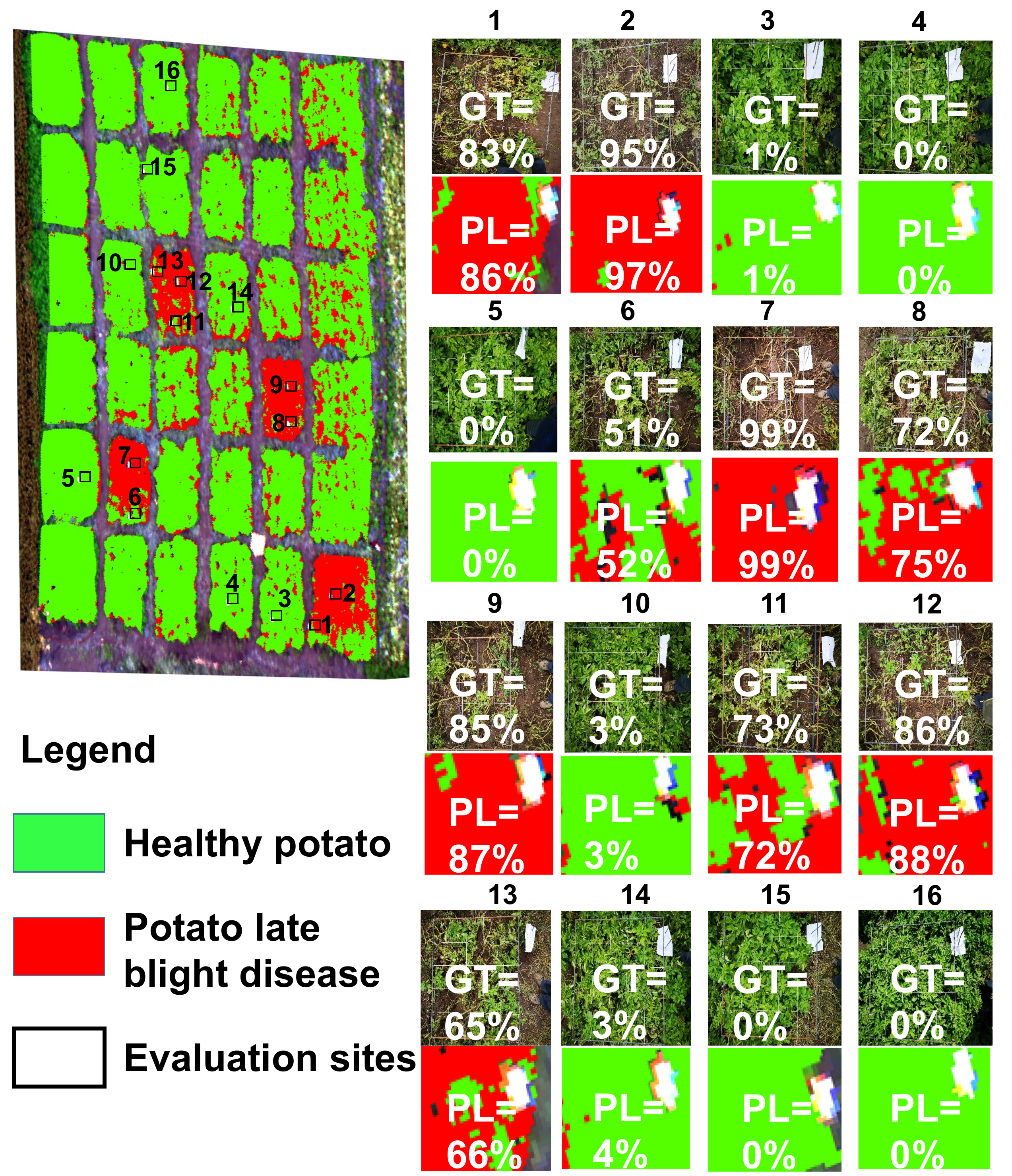}   
\caption{The patch-scale test for the classification maps of the healthy potato and the potato late blight disease in experimental site 2. Here, the example patches on the right side illustrate the accuracy comparison between the ground truth (GT) investigations and the predicted levels (PL) of the late blight disease. Each value inside the patch represents the disease ratio.  }  
\label{fig:class_map2}  
\end{figure}

Fig. \ref{fig:class_map1} and Fig. \ref{fig:class_map2} show the patch-scale test for the classification maps of healthy potato and potato late blight disease overlaying on the UAV HSI in experimental site 1 and site 2, respectively. For the experimental site 1,  $9$ patches with $1m \times 1m$ size are ground truth data. Our results illustrate that, the average differences of disease ratio within the patches between the ground truth data and the classification map is $2.6\%$. The maximum difference occurring in the patch $8$ is $5\%$. For the experimental site 2, there are $16$ $1m \times 1m$ ground truth patches. Our findings suggest that the average differences of disease ratio within the patches between the ground truth patches and the patches from the classification map is $1\%$, and the maximum difference occurring in the patch $1$ is $3\%$. \par

\begin{table}[]
\caption{The confusion matrix of the proposed model on the pixel scale detection and discrimination of potato late blight disease. Here, U is the User's accuracy, P is the Producer's accuracy. }
\label{table:confuse}
\centering
\resizebox{10 cm}{!}{
\begin{tabular}{cccccccc}
\toprule
                    & Healthy potato & Late blight disease & Soil & Background & UA(\%) & OA(\%) & Kappa \\ \midrule
Healthy potato      & 81             & 1                   & 0    & 0          & 98.8            &        &       \\
Late blight disease & 2              & 82                  & 0    & 0          & 97.6            & 97     & 0.812 \\
Soil                & 0              & 2                   & 19   & 0          & 90.5            &        &       \\
Background          & 0              & 0                   & 1    & 12         & 92.3            &        &       \\
PA(\%)               & 97.5           & 96.3                & 94.7 & 100        &                 &        &       \\ \bottomrule
\end{tabular}
}
\end{table}

\section{Discussion}
\label{sec:4}
In this paper, we propose a CropdocNet for learning the late blight disease associated hierarchical structure information from the UAV HSI data, and providing more accurate crop disease diagnosis at the farm scale. Unlike the traditional spectral-spatial scalar features used in the existing machine learning/deep learning approaches,  our proposed method considers the hierarchical structure of the late blight disease associated spectral-spatial characteristics, to represent rotation invariance of the late blight disease under the complicated field conditions. This is achieved by introducing the capsule layers into our network architecture to construct the hierarchical structure learning mechanism of the late blight disease-specific characteristics. 

The hierarchical structure of the spectral-spatial information extracted from HSI have been proven effective for representing the invariance of the target entities on HSI \cite{shi2021biologically}. To investigate how the potato late blight disease detection can benefit from using the hierarchical vector features on the UAV based HSI, we have compared the proposed model with three typical machine learning models considering only the spectral or spatial scalar features. The results illustrate that the proposed model outperforms the traditional models in terms of overall accuracy, average accuracy, sensitivity and specificity. In addition, the classification differences between the proposed model and the existing models are statistically significance.  \par 

To trade off the accuracy and computing efficiency, the effects of the depth of the convolutional filters are investigated. Our findings suggest that there is no obvious improvement in accuracy when the depth of 1-D convolutional kernels $K^{(1)} = 128$ and the depth of 3-D convolutional kernels $K^{(2)} = 64$. We also find that, by using the multi-scale capsule units ($K^{(3)} = 32$), the model performance on HSI-based potato late blight disease detection could be improved. \par

\begin{figure}[]   
\centering  
\includegraphics[width=10.5 cm]{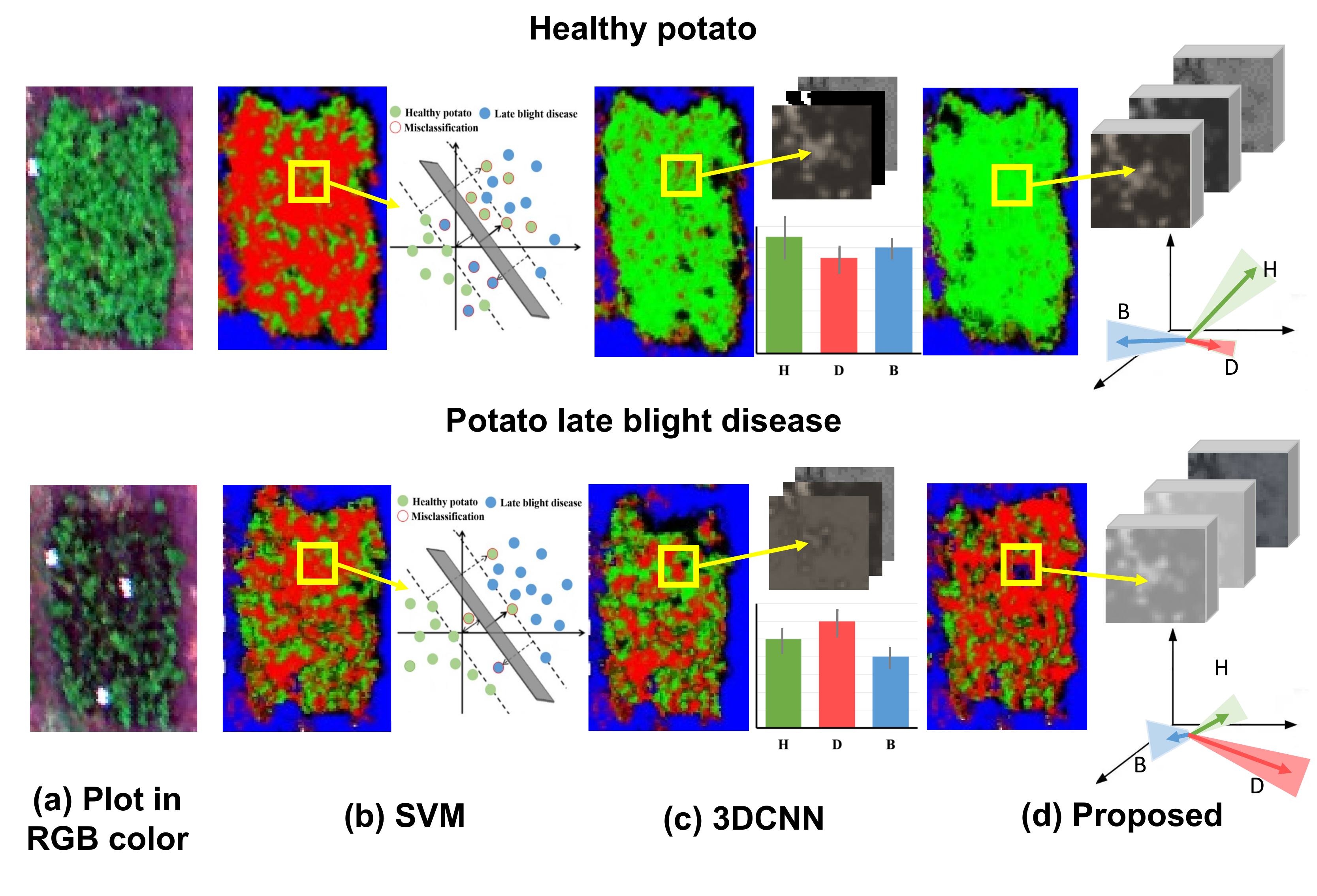}   
\caption{The visualized feature space and the mapping results of the healthy and diseased plots based on the different machine learning/deep learning methods: a) the original RGB image for the healthy potato (H) and diseased potato (D), and background (B). b) the classification results and the visualized spectral feature space of SVM, c) the classification results and the averages and the standard deviations of the activated high-level spectral-spatial features of 3DCNN, and d) the classification results and the visualized hierarchical capsule feature space of the proposed CropdocNet.  }  
\label{fig:visual}  
\end{figure}

To further visually demonstrate the benefit of using hierarchical vector features in the proposed CropdocNet, we have compared the visualized feature space and the mapping results of the healthy (see the first row of Fig. \ref{fig:visual}) and diseased plots (see the second row of Fig. \ref{fig:visual}) from three models: SVM, 3DCNN, and the proposed CropdocNet. Our quantitative assessment reveals that the accuracy of the potato late blight disease plots is $76.8\%$, $83.2\%$, and $94.2\%$ for SVM, 3DCNN, and CropdocNet, respectively. Specifically, for the SVM-based model which only maps the spectral information into the feature space, a total of $81\%$ areas in the healthy plots are misclassified as the potato late blight disease (see the left subgraph of Fig. \ref{fig:visual}b). And the feature space of the samples in the yellow frame, as shown in the right subgraph of Fig. \ref{fig:visual}b, explains the reason for these misclassifications. Thus, there is no cluster characteristics can be observed between the spectral features in the SVM-based feature space, indicating that the inter-class spectral variances are not significant between the SVM decision hyperplane.

In contrast, the spectral-spatial information based on 3DCNN (\ref{fig:visual}c) performs better than the SVM-based model. However, looking at the edge of the plots, there are obvious misclassifications. The right subgraph of Fig. \ref{fig:visual}c, reveals the averages and the standard deviations of the activated high-level features of the samples within the yellow frame. It is worth noting that, for the healthy potato (the first row of Fig. \ref{fig:visual}c), the average values of the activated joint spectral-spatial features for different classes are quite close, and the standard deviations are relatively high, illustrating that the inter-class distance between the healthy potato and the potato late blight disease are not significant in the features space. The similar results can be found in the late blight disease (see second row of Fig. \ref{fig:visual}c). Thus, no significant inter-class separability can be represented in the joint spectral-spatial feature space owning to the mixed spectral-spatial signatures of plant and background. \par

In comparison, the hierarchical vector features-based CropdocNet model provides more accurate classification because the hierarchical structural capsule features can express the various spectral-spatial characteristics of the target entities. For example, the white panels in the diseased plot (see the second row of of Fig. \ref{fig:visual}d) are successfully classified as the background. The right subgraphs of Fig. \ref{fig:visual}d demonstrate the average, direction, and standard deviations of the activated hierarchical capsule features of the samples within the yellow frame. It's noteworthy that the average length and direction of the activated features for different classes are quite different, and the standard deviations (see the shadow under the arrows) do not overlap with each other. These results fully demonstrate the significant clustering of each class in the hierarchical capsule feature space. In conclusion, benefiting from the hierarchical capsule features, the proposed CropdocNet performs better on potato late blight disease detection than the existing spectral based or spectral-spatial based models. \par

\section{Conclusions}
\label{sec:5}
In this study, a novel CropdocNet is proposed for extracting the spectral-spatial hierarchical structure of the late blight disease, and automatically detecting the disease from the UAV HSI imagery. The innovation of the CropdocNet is the hierarchical network architecture that integrates the spectral-spatial scalar features into the hierarchical vector features for representing the rotation invariance of the potato late blight disease in the complicated filed conditions. The model has been tested and evaluated on the real field data, and compared with the existing machine/deep learning models. The experimental findings demonstrate that the proposed model is able to significantly improve the accuracy of the potato late blight disease on the HSI data. The future work will be to validate the proposed model on more UAV-based HSI data with various potato growth stages and different diseases. 

\section{Acknowledge}
This research is supported BBSRC BBSRC (BB/S020969/1), (BB/R019983/1).

\bibliographystyle{unsrtnat}
\bibliography{Ref}  






\end{document}